\documentclass[journal]{IEEEtran}

\usepackage{amsmath}
\usepackage{amssymb}
\usepackage{graphicx}
\usepackage{multirow}
\usepackage{url}
\usepackage{float} 
\usepackage{color}
\usepackage{layouts}

\newcommand{\argmin}{\text{argmin}}

\definecolor{myred}{rgb}{.8,.0,.0}
\definecolor{mynormal}{rgb}{.0,.0,.0}

\newcommand{\added}[1]{\textcolor{mynormal}{#1}}

\newcommand{\linelabel}{}

\begin{document}

\title{Transfer learning for multi-center classification of chronic obstructive pulmonary disease}


\author{Veronika~Cheplygina,
        Isabel Pino Pe{\~n}a,
        Jesper Holst Pedersen,
        David A. Lynch,
        Lauge S{\o}rensen,
        and~Marleen~de~Bruijne
\thanks{V. Cheplygina was with the Biomedical Imaging Group Rotterdam, Departments of Medical Informatics and Radiology of the Erasmus MC - University Medical Center Rotterdam, Rotterdam, The Netherlands when this work was performed. She is now with the Medical Image Analysis group, Eindhoven University of Technology. E-mail: v.cheplygina@tue.nl}%
\thanks{I. Pino Pe{\~n}a is with the Department of Health Science and Technology, Aalborg University, Aalborg, Denmark.}%
\thanks{J. H. Pedersen is with the Department of Thoracic Surgery, Rigshospitalet, University of Copenhagen, Copenhagen, Denmark.}%
\thanks{D. A. Lynch is with the Department of Radiology, National Jewish Health, Denver, Colorado, United States of America.}%
\thanks{L. S{\o}rensen and M. de Bruijne are with the Image Section, Department of Computer Science, University of Copenhagen,  Copenhagen, Denmark.}%
\thanks{M. de Bruijne is also with the Biomedical Imaging Group Rotterdam, Departments of Medical Informatics and Radiology of the Erasmus MC - University Medical Center Rotterdam, Rotterdam, Netherlands. E-mail: marleen.debruijne@erasmusmc.nl}
}

\maketitle

\begin{abstract}
Chronic obstructive pulmonary disease (COPD) is a lung disease which can be quantified using chest computed tomography (CT) scans. Recent studies have shown that COPD can be automatically diagnosed using weakly supervised learning of intensity and texture distributions. However, up till now such classifiers have only been evaluated on scans from a single domain, and it is unclear whether they would generalize across domains, such as different scanners or scanning protocols. To address this problem, we investigate classification of COPD
in a multi-center dataset with a total of 803 scans from three different centers, four different scanners, with heterogenous subject distributions. Our method is based on Gaussian texture features, and a weighted logistic classifier, which increases the weights of samples similar to the test data. We show that Gaussian texture features outperform intensity features previously used in multi-center classification tasks. We also show that a weighting strategy based on a classifier that is trained to discriminate between scans from different domains, can further improve the results. To encourage further research into transfer learning methods for classification of COPD, upon acceptance of the paper we will release two feature datasets used in this study on \url{http://bigr.nl/research/projects/copd}. 
\end{abstract}


\section{Introduction}

Chronic obstructive pulmonary disease (COPD) is characterized by chronic inflammation of the lung airways and emphysema, i.e., degradation of lung tissue~\cite{pauwels2001global}. Emphysema can be visually assessed in vivo using chest computed tomography (CT) scans, however, to overcome limitations of visual assessment, automatic quantification of emphysema has been explored~\cite{park2008texture,sorensen2010quantitative,mendoza2012emphysema,sorensen2012texture,cheplygina2014classification}. Several of these methods rely on supervised learning and require manually annotated regions of interest (ROIs)~\cite{park2008texture,sorensen2010quantitative,mendoza2012emphysema}, while other approaches using multiple instance learning (MIL) only require patient-level labels indicating overall disease status~\cite{sorensen2012texture,cheplygina2014classification}. In this work we address this weakly-supervised classification setting, i.e., the scans are only labeled as belonging to a COPD or non-COPD subject, and no information on ROI level is available. \linelabel{categorization} \added{The problem can be seen as a categorization (assign scan to a COPD or non-COPD category) problem or as a detection (detect whether COPD is present in the scan) problem; to be consistent with machine learning terminology we refer to this problem as ``classification''. Although we do not focus on quantification (quantifying the grade of COPD in the scan), we discuss how our classification method can be adapted for this purpose}.  

A challenge for classification of COPD in practice is that the training data may not be representative of the test data, i.e. the distributions of the training and the test data are different. This can happen if the data originates from different \emph{domains}, such as different subject groups, scanners, or scanning protocols. One approach to overcome this problem is to search for features that are robust to such variability. For example, in a multi-cohort study with different CT scanners~\cite{mendoza2012emphysema}, the authors compare intensity distribution features to local binary pattern (LBP) texture features, and suggest that intensity might be more effective in multi-scanner situations. 


\added{Another way to explicitly address the differences in the distributions of the training and test data is called \emph{transfer learning}~\cite{pan2010survey} or domain adaptation. \linelabel{thesedifferences} These differences can be caused by different marginal distributions $p(\mathbf{x})$, different labeling functions $p(y|\mathbf{x})$, or even different feature and label spaces. In this work the $\mathbf{x}$'s are the feature vectors describing the appearance of the lungs, and the $y$'s are the categories the subjects belong to. Changes in subject groups, scanners and scanning protocols, can affect the distributions $p(\mathbf{x})$, such as ``this dataset has lower intensities'', $p(y)$, such as ``this dataset has more subjects with COPD'' and/or $p(y|\mathbf{x})$, such as ``in this dataset this appearance corresponds to a different category''.} 

Based on which distributions are the same, and which distributions are different, different transfer learning scenarios can be distinguished. One of these scenarios is transductive transfer learning, where labeled training data (or source data), as well as unlabeled test data (or target data), are assumed to be available. This is the scenario we investigate. 

\added{According to ~\cite{pan2010survey}}, transfer learning methods can be divided into instance-transfer, feature-transfer, parameter-transfer and relational-knowledge-transfer approaches. This paper presents an instance-transfer approach, but we briefly discuss instance-transfer and feature-transfer, \added{which are most abundant in medical imaging}, in order to contrast our work from the literature. In short, feature-transfer approaches aim to find features which are good for classification, possibly in a different classification problem. In contrast, instance-transfer methods aim to select source samples which help the classifier to generalize well. \added{One intuitive instance-transfer approach is called ``importance weighting''}~\cite{shimodaira2000improving,huang2006correcting,gretton2009covariate}, i.e., assigning weights to the source samples, based on their similarity to the unlabeled target samples, and subsequently training a weighted classifier. This strategy assumes that only the marginal distributions \added{$p(\mathbf{x})$} are different, and that the labeling functions \added{$p(y|\mathbf{x})$} are the same. However, in practice, importance weighting can also be beneficial in cases where the labeling functions are different~\cite{opbroek2015weighting}. 

Transfer learning techniques are relatively new in the medical imaging domain, and have shown to be successful in several applications, such as classification of Alzheimer's disease~\cite{cheng2015multimodal,guerrero2014manifold} and segmentation of magnetic resonance (MR) images~\cite{opbroek2015weighting,opbroek2014transfer} and microscopy images~\cite{becker2014domain,ablavsky2012transfer}. In chest CT scans, transfer learning has been used for classification of different abnormalities in lung tissue~\cite{schlegl2014unsupervised,shin2016deep}. However, these approaches focus on feature-transfer between datasets, possibly even from non-medical datasets, while we investigate an instance transfer approach which focuses on differences between data acquired at different sites. To the best of our knowledge, our work is the first to investigate the use of transfer learning for classification of COPD. 

The contributions of this paper are twofold. Our first contribution is a comparison of different types of intensity- and texture-based features for the task of classifying COPD in chest CT scans, to assess the features' robustness across scanners. The second contribution is a proposed approach which combines transfer learning with a weakly-supervised classifier. To this end, we investigate three different weighting strategies. We use four datasets, which differ with respect to the subject group, site of collection, scanners and scanning protocols used. Furthermore, we publicly release two feature datasets used in this study to further the progress in transfer learning in classification of COPD and in medical image analysis in general.

\section{Methods}\label{sec:methods}

\begin{figure*}
    \centering
    \includegraphics[width=0.9\textwidth]{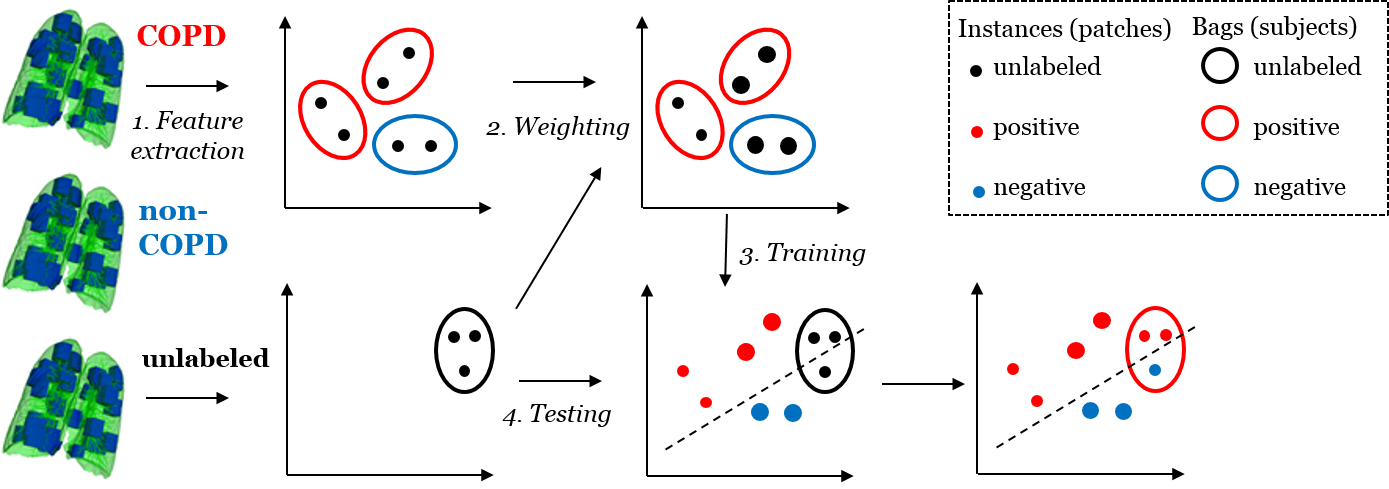}
    \caption{Overview of the procedure. \added{Step 1 is to represent each scan as a bag of instances (feature vectors). The bag is labeled as COPD (+1) or non-COPD (-1). Step 2 is to weight the training bags by their similarity to the unlabeled test scan. Step 3 is to use the weighted bags to train a classifier. In step 4 this classifier is used to classify the test instances. The instance labels are combined into an overall label for the scan, in this case COPD (+1).}}
    \label{fig:procedure}
\end{figure*}

Following S{\o}rensen et al.~\cite{sorensen2012texture}, we represent each chest CT image by a set of 3D ROIs. Each ROI is represented by a feature vector describing the intensity and/or texture distribution in that ROI. In order to classify each individual test scan, we assign weights to the training scans based on their similarity to the test scan, and subsequently train a weighted multiple instance classifier. The procedure is illustrated in Fig.~\ref{fig:procedure}.

\subsection{Notation and Feature Representation} \label{sec:representation}
Each scan is represented by a bag $X_i  = \{\mathbf{x}_{ij}| j=1,...,n_i\} \subset \mathbb{R}^m$ of $n_i$ instances (ROIs), where the $j$-th instance or ROI is described by a $m$-dimensional feature vector $\mathbf{x}_{ij}$. The bags have labels $y_i \in \{+1, -1\}$, in our case COPD and non-COPD, but the instances are unlabeled: the problem is thus called weakly supervised. The bags originate from two different datasets: training (source) and test (target) data. We will denote bags and instances from the source data by $X$ and $\mathbf{x}$, bags and instances from the target data are denoted by $Z$ and $\mathbf{z}$. \added{Both the distributions $p(X)$ and $p(Z)$, and the distributions $p(y|X)$ and $p(y|Z)$ may be different from each other.} 

We represent each CT scan by a bag of 50 possibly overlapping, volumetric ROIs of size $41\times 41 \times 41$ voxels, extracted at random locations inside the lung mask. The lung masks were obtained prior to this study. For three datasets (DLCST and both COPDGene datasets), the lung masks were obtained with a region-growing algorithm and postprocessing step used in \cite{lo2010vessel}, and for one dataset (Frederikshavn) with a method based on multi-atlas registration and graph cuts, similar to~\cite{korsager2015use}. 

We use Gaussian scale space (GSS) features, which capture the image texture, to represent the ROIs. Each image is first convolved with a Gaussian function at scale $\sigma$ using normalized convolution within the lung mask. We use four different scales, \{0.6, 1.2, 2.4, 4.8\} mm, and compute eight filters: smoothed image, gradient magnitude, Laplacian of Gaussian, three eigenvalues of the Hessian, Gaussian curvature and eigen magnitude. The filtered outputs are summarized with histograms, where adaptive binning~\cite{ojala1996comparative} is used to best describe the data while reducing dimensionality. We quantize the output of each filter into ten bins, where the bin edges used for adaptive binning (i.e. volume in each bin must be equal) of all datasets have been determined on an independent sample from one of the datasets (DLCST). This leads to $8 \times 4 \times 10 = 320$ features in total.

\subsection{Classifier}\label{sec:classifier}

To learn with weakly labeled scans, we use a MIL classifier. \added{In particular we use an approach we refer to SimpleMIL, which could be also seen as a naive MIL classifier. The name SimpleMIL was first used in~\cite{cheplygina2014classification}, but the approach is older, and is often used as a baseline, even if MIL is not mentioned, for example in~\cite{loog2004static}. SimpleMIL propagates the training bag labels to the training instances, and trains a supervised classifier.} We use a weighted logistic classifier $\mathbf{w}^*$, defined as follows: 

\begin{equation}\label{eq:logistic}
\mathbf{w^*} = \argmin_{\mathbf{w}} ( \sum_{(\mathbf{x}_{ij},y_{ij})} s_{ij} L(\mathbf{w}, \mathbf{x}_{ij}, y_{ij}) + \lambda ||\mathbf{w}||_2^2),
\end{equation}
where $\mathbf{w}$ is a vector of $m$ feature coefficients (we drop the intercept for ease of notation), the loss is defined as $L(\mathbf{w},\mathbf{x},y) = \frac{1}{\ln{2}} \ln{(1+\exp{(-y \mathbf{w}^{\intercal} \mathbf{x}}}))$, $\lambda$ is a regularization term controlling the complexity of the weight vector, and $s_{ij}$ is the importance weight associated with the $j$-th instance from the $i$-th bag (see Section \ref{sec:weighting}). 

To make sure that the total effect of weights is the same across different weighting strategies, before training the classifier we multiply the weights by $N/\sum_{i,j} s_{ij}$, such that the sum of the weights is equal to the number of training instances $N$. 

\added{When a test bag $Z_i$ is presented to the classifier, $\mathbf{w}^*$ is used to obtain posterior probabilities $p(y_{ij}=+1 | \mathbf{z}_{ij})$ and $p(y_{ij}=-1  | \mathbf{z}_{ij})$ for each instance $ \mathbf{z}_{ij}$.} A posterior probability for the test bag is obtained by combining the instance posteriors. Here we apply the average rule, 
\begin{equation}\label{eq:average}
\frac{p(y_i=+1|Z_i)}{p(y_i=-1|Z_i)} = \frac{1}{n_i} \sum_{j=1}^{n_i} \frac{ p(y_{ij}=+1|\mathbf{z}_{ij})} {p(y_{ij}=-1 | \mathbf{z}_{ij})},
\end{equation}
which assumes that all instances contribute to the bag label. \added{In other words, on average, the instances should be classified as positive - it is not sufficient if only a few instances are positive. This is consistent with the observation that COPD is not a localized disease, but more spread out throughout the lung. A combining strategy similar to the average rule (thresholding the posteriors and then combining the decisions, rather than combining the probabilities as we do here) was used for classification of COPD in \cite{park2008texture}. Furthermore, the average rule has been used in other medical imaging applications~\cite{loog2004static,quellec2012multiple}}. \added{Other assumptions, where only a single positive instance is needed for a positive bag, are also possible and will be discussed in Section \ref{sec:discussion}. }.

Despite the simplicity of this approach, this strategy has achieved good results in previous experiments on weakly-labeled single-domain chest CT data~\cite{sorensen2012texture,cheplygina2014classification}. In~\cite{cheplygina2014classification}, this method was used with a logistic and a nearest neighbor classifier, and the logistic classifier achieved the best performance. 


\subsection{Instance Weighting}\label{sec:weighting}
We estimate the weights of the source bags with three different weight measures: 
\begin{itemize}
    \item using the distance from the source bags to the target bag,
    \item using the distance from the target bag to the source bags,
    \item using the estimated probability of the source bag belonging to the target class
\end{itemize}

In the traditional instance weighting approach, the weights are assigned to instances $\mathbf{x}$, which are considered independent. However, for MIL, this is not the most intuitive approach, \added{since the different instances $\mathbf{x}_{ij}$ within the same bag $X_i$ are expected to be correlated}. Therefore, rather than finding similar instances in the training data, we are more interested in similar bags. Because we want to assign the weights on bag-level, in what follows we describe how to obtain a weight $s_i$ for each bag $X_i$. In training the SimpleMIL classifier, however, each instance is associated with a weight equal to the bag weight, i.e. $s_{ij} = s_i$. 

\added{By weighting the training samples, we aim for the weighted distribution $(p\mathbf{x})$ to become more similar to $p(\mathbf{z})$, and thus for the trained classifier to provide more accurate estimates $p(y|\mathbf{z})$.}

\subsubsection{Source to target weights}\label{sec:s2t}

The first approach is based on a bag distance between the source bag, and the target bag. We use weights that are inversely proportional to the source-to-target (\emph{s2t}) distance of source bag $X_i$ to a target bag $Z$. In converting the distances to weights, we scale the weights to the interval [0,1], which assumes that there are always relevant and irrelevant source samples. The \emph{s2t} weights are then defined as follows:

\begin{equation}\label{eq:w_s2t}
s^{s2t}_{i} = \frac{d^{s2t}_{max}-d^{s2t}_{i}}{d^{s2t}_{max}-d^{s2t}_{min}}
\end{equation}
where
\begin{equation}\label{eq:dist_s2t}
d^{s2t}_{i} = \frac{1}{|X_i|}\sum_{\mathbf{x}_{ij} \in X_i} \min_{\mathbf{z}_k \in Z} ||\mathbf{x}_{ij} - \mathbf{z}_k ||^2 .
\end{equation}
and $d_{max} = \max_{i} d_{i}$ and $d_{min} = \min_{i} d_{i}$ are the maximum and minimum bag distances found in the training set. 

In other words, for each instance in the source bag, we find its nearest neighbor in the target bag $Z$, and average the nearest neighbor distances. A divergence measure that is analogous to this distance has been successfully used in previous works on transfer learning in medical image analysis~\cite{opbroek2015weighting,opbroek2014transfer}. The distance we propose is more efficient to compute, and has shown to be robust in high-dimensional situations~\cite{cheplygina2016asymmetric} \added{(and references therein)} than related divergences.

\subsubsection{Target to source weights}\label{sec:t2s}

The matching of instances with their nearest neighbors makes the bag distance asymmetric. In previous work on medical imaging such asymmetry was important for classification performance~\cite{cheplygina2016asymmetric}. The rationale is that for a test scan with unusual ROIs (i.e., outliers in feature space), we want to ensure that these outliers influence the training weights as much as possible. However, with the \emph{s2t} distance, it is possible that the test outliers do not participate in the weighting process at all. Therefore we also examine weights based on the counterpart of the source-to-target distance, i.e. the target-to-source (\emph{t2s}) distance:

\begin{equation}\label{eq:w_t2s}
s^{t2s}_{i} = \frac{d^{t2s}_{max}-d^{t2s}_{i}}{d^{t2s}_{max}-d^{t2s}_{min}}
\end{equation}
where
\begin{equation}\label{eq:dist_t2s}
d^{t2s}_{i} = \frac{1}{|Z|}\sum_{\mathbf{z}_k \in Z} \min_{\mathbf{x}_{ij} \in X_i} ||\mathbf{x}_{ij} - \mathbf{z}_k ||^2 
\end{equation}

and $d^{t2s}_{max}$ and $d^{t2s}_{min}$ are defined analogously to $d^{s2t}_{max}$ and $d^{s2t}_{min}$. 

Note that we can only use the \emph{t2s} distance for weighting because we are computing bag distances. If we would weight the training instances independently, some of the training instances might not get matched with target instances, and therefore might not receive a weight.

\subsubsection{Logistic weights}

The last weighting approach is based on how well a logistic classifier $\mathbf{w}^s$, which models posterior probabilities, can separate the source and target data.  That is, all the instances in the source data are labeled as class -1, and samples in the target data are labeled as class 1, and the classifier $\mathbf{w}^s$ is trained on these two classes. The source samples are then evaluated by the classifier to obtain their probabilities of belonging to the target class $p(y=1 | \mathbf{x}_{ij}) = \exp{(-\mathbf{w}^{s \intercal}\mathbf{x}_{ij})} /  \sum_{y_{ij} 
in \{-1,+1\}}\exp{(-y_{ij}\mathbf{w}^{s \intercal}\mathbf{x}_{ij})}$. For a training bag, we therefore have the following: 

\begin{equation}\label{eq:dist_log}
s^{log}_{i} = \frac{1}{|X_i|}\sum_{\mathbf{x}_{ij} \in X_i} \frac{\exp{(-\mathbf{w}^{s \intercal}\mathbf{x}_{ij})}}{\sum_{y_{i} \in \{-1, +1\}} \exp{(-y_{i} \mathbf{w}^{s \intercal}\mathbf{x}_{ij})}}
\end{equation}

This approach is common in transfer learning literature in the field of machine learning~\cite{kouw2015feature}, and, in the infinite-sample case and no change in labeling function, has shown to be equivalent to a classifier trained on the source samples~\cite{shimodaira2000improving}. In medical image analysis, this approach has been used for segmentation of tumors in brain MR images~\cite{goetz2015dalsa} for a domain adaptation setting where only the sampling of the training and test data is different. 


\section{Experiments}

\subsection{Data} \label{sec:data}
We use four datasets from different scanners in the experiments (Table~\ref{tab:details}). The first dataset consists of 600 baseline inspiratory chest CT scans from the Danish Lung Cancer Screening Trial  ~\cite{pedersen2009danish}. The second (120 inspiratory scans) and third (67 inspiratory scans) datasets consist of subjects from the COPDGene study~\cite{regan2011genetic}, both acquired at the National Jewish Center in Denver, Colorado. The fourth dataset (16 scans) consists of subjects with respiratory problems referred to the out-patient clinic of the Frederikshavn hospital in Denmark. \added{We refer to these datasets as DLCST, COPDGene1, COPDGene2 and Frederikshavn throughout the paper.} 

 All scans are acquired at full inspiration, and the COPD diagnosis is determined according to the Global Initiative for Chronic Obstructive Lung Disease (GOLD) criteria~\cite{vestbo2013global}, i.e., $FEV_1 / FVC < 0.7$. \linelabel{asinprevious} \added{As in previous work by the authors~\cite{sorensen2012texture,cheplygina2014classification} and in other literature~\cite{phillips2012machine,amaral2012machine} where COPD categorization is addressed with machine learning methods (but without using imaging data), we consider a binary classification problem. In other words, we treat subjects with GOLD grade 0 as the non-COPD class, and subjects with GOLD grades between 1 and 4 as the COPD class}. 

We consider DLCST, COPDGene1 and COPDGene2 both as source data and as target data, and Frederikshavn only as target data, due to its small size. 

\begin{table*}[ht]
\begin{center}
\begin{tabular}{l l l l l l l l l  }

Dataset & Subjects & Age & GOLD & Smoking & Scanner & Resolution (mm) & Exposure & Reconstruction \\
  &    &   &  (1/2/3/4) & (c/f/n) &  &   &   &   \\

\hline

DLCST & 300 + & 59 [50, 71] & 69/28/2/0 & 77/23/0 & Philips & 0.72$\times$0.72$\times$1 to  & 40 mAs & Philips D \\
      & 300 - & 57 [49, 69] & & 74/26/0 & 16 rows Mx 8000 & 0.78$\times$0.78$\times$1 &   &  hard  \\
            
           
COPDGene1    & 74 + & 64 [45, 80] &  21/18/19/16    &  17/57/0  &  Siemens  & 0.65$\times$0.65$\times$0.75 &  200 mAs & B45f sharp  \\
             & 46 - & 59 [45, 78] &     &  23/20/3    &  Definition &  &  &   \\ 
               
COPDGene2  & 42 + & 65 [45, 78] & 9/13/7/13   & 12/30/0   &   Siemens      &  0.65$\times$0.65$\times$0.75 &  200 mAs & B45f sharp  \\
           & 25 - & 60 [47, 78] &   & 9/11/5   &   Definition AS+ &  &  &   \\ 

Frederikshavn  & 8 + & 66 [48, 77] & 1/3/3/1  & 1/7/0 & Siemens  &  0.58$\times$0.58$\times$0.6  &  95 mAs   &  I70f very sharp \\
             & 8 - & 56 [25, 73] &          & 1/2/5 & Definition Flash     &           &           \\

\end{tabular}
\caption{Details of datasets. For subjects, + = COPD, - = non-COPD. Ages reported as mean [min, max], rounded to nearest integer. GOLD refers to the COPD grade as defined by the Global Initiative for Chronic Obstructive Lung Disease. For smoking status, c=current, f=former, n=never.}
\label{tab:details}
\end{center}
\end{table*}

\subsection{Feature Datasets}

In the proposed approach, each ROI is represented by GSS as described in Section~\ref{sec:representation}, resulting in a feature vector with 320 dimensions. We compare our method with intensity features based on kernel density estimation (KDE) used in~\cite{mendoza2012emphysema}. We use 256 bins in order for the dimensionality to be comparable to the Gaussian features. To focus on the more informative part of the intensities, we apply the KDE to the range [-1100HU, -600HU]. We originally used a larger range and 4096 bins, following correspondence with the authors of ~\cite{mendoza2012emphysema}. However, this gave poor results in preliminary experiments on DLCST and Frederikshavn data. We concluded that the classifier suffered from overfitting, and adapted the range and dimensionality to produce reasonable results for those two datasets. 

Furthermore, we compare our feature set to two of its subsets: a subset with 40 features describing the intensity of the scan at different scales (GSS-i), and its complement with 280 features describing with derivatives only, thus more describing the texture (GSS-t). These comparisons will allow us to evaluate whether it is the intensity information that is responsible for differences between GSS and KDE, or the particular choice of implementation used in KDE.

\subsection{Classifiers without Transfer Learning}

We first use SimpleMIL with a logistic classifier without any weighting. We train classifiers on each of the three source datasets (DLCST, COPDGene1 and COPDGene2). We then apply the trained classifiers to the four target datasets (DLCST, COPDGene1, COPDGene2 and Frederikshavn). When the source and target datasets are the same, this experiment is performed in a leave-one-scan-out procedure. 

The logistic classifier has only one free parameter, the regularization parameter $\lambda$. For both $\mathbf{w}^*$ (the SimpleMIL classifier) and $\mathbf{w}^s$ (the classifier used to determine the logistic weights) we fix $\lambda = 1$, because in preliminary experiments choosing other values did not have a large effect on the results.

\subsection{Classifiers with Transfer Learning}

We then use SimpleMIL with a weighted logistic classifier. For each of the nine combinations of source and different-domain target datasets, we perform a leave-one-image-out procedure. For each target image, we determine the weights using three different methods: \emph{s2t}, \emph{t2s} and logistic. We then train the weighted classifiers and evaluate them on the target image. Again, the regularization parameter $\lambda$ is fixed to 1.

\subsection{Evaluation}

The evaluation metric is the area under the receiver-operating characteristic curve (ROC), or AUC. \linelabel{significant} We test for significant differences using the DeLong test for ROC curves~\cite{delong1988comparing}.

To summarize results over nine pairs of source and target data, we also rank the different weighting methods and different feature methods, and report the average ranks. To assess significance, we perform a Friedman/Nemenyi test~\cite{demvsar2006statistical} at the $p=0.05$ level. This test first checks whether there are any significant differences between the ranks, and if so, determines the minimum difference in ranks (or critical difference) required for any two individual differences to be significant. For nine pairs of datasets and four methods, the critical difference is 1.56.

\section{Results}

\subsection{Performance without Transfer Learning}

\begin{figure*}
    \centering
    \includegraphics[width=0.4\textwidth]{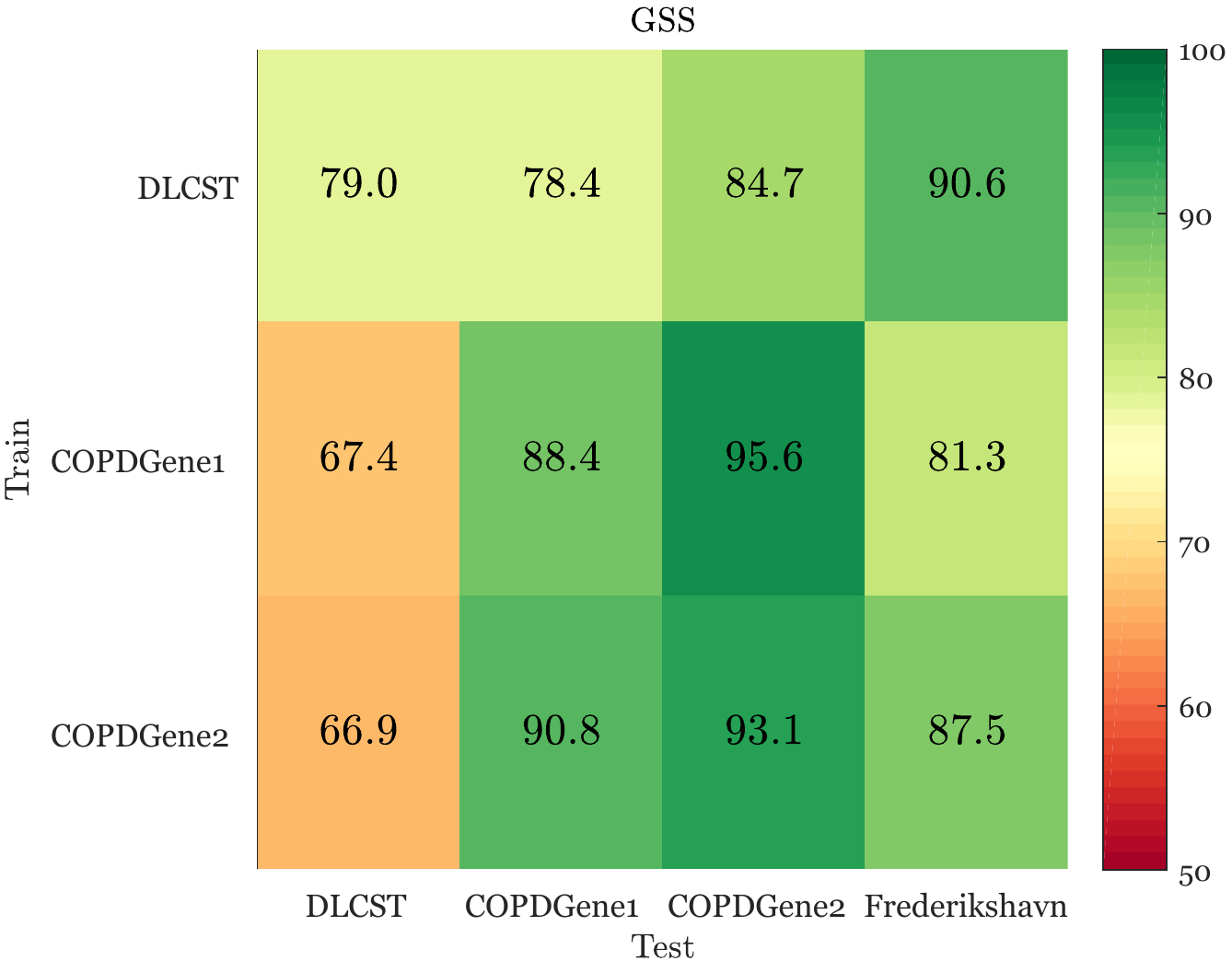}
    \includegraphics[width=0.4\textwidth]{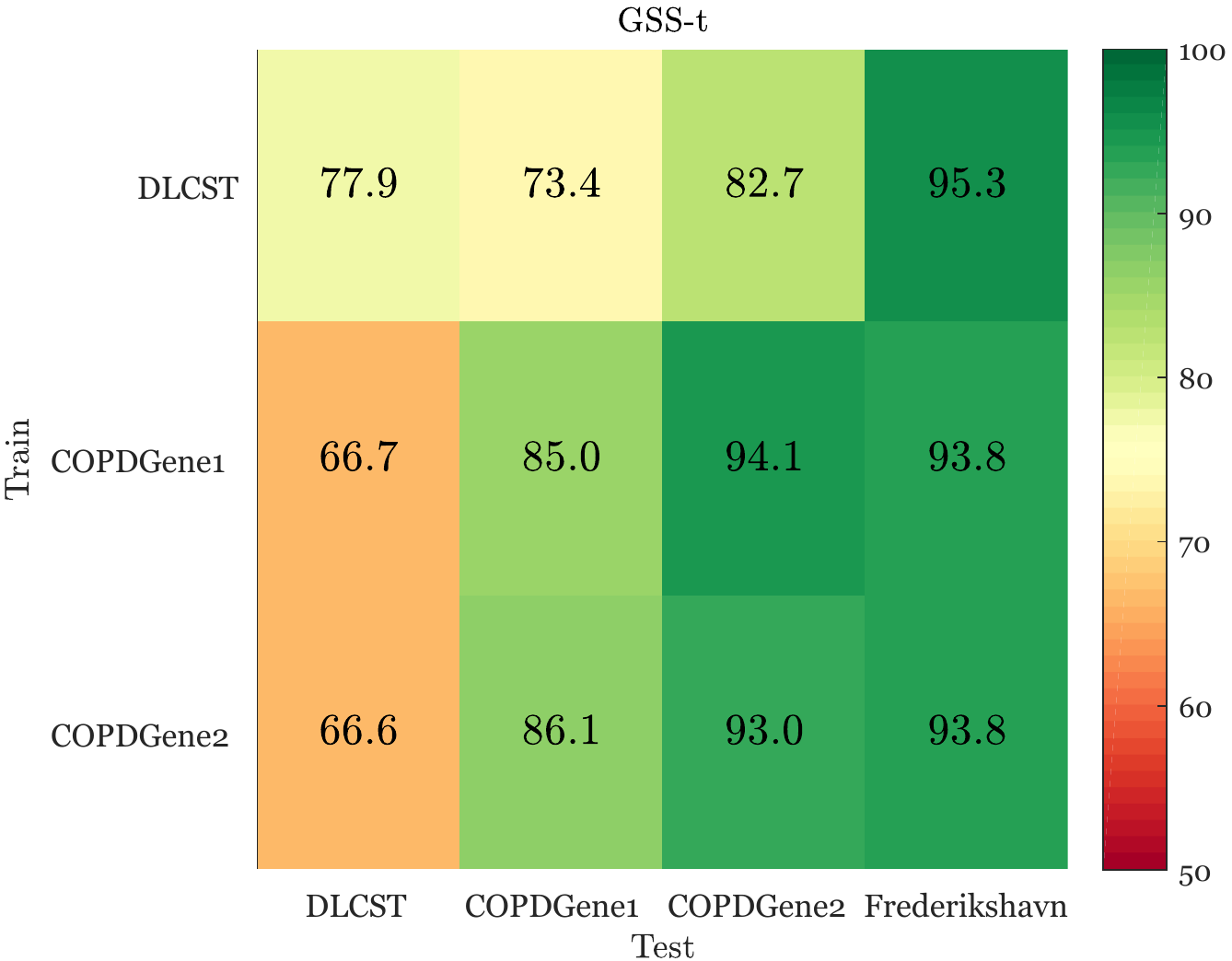}
    
    \includegraphics[width=0.4\textwidth]{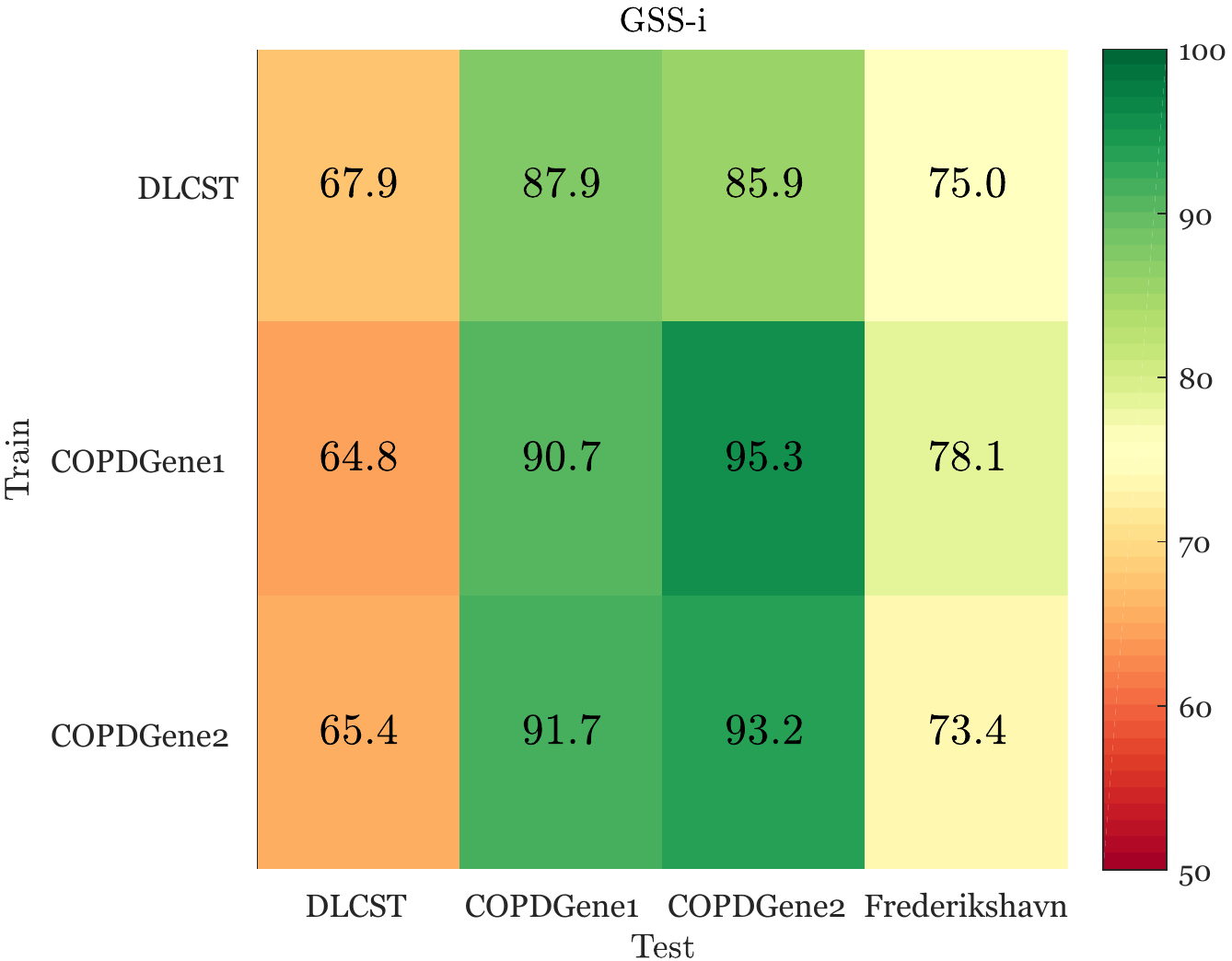}
    \includegraphics[width=0.4\textwidth]{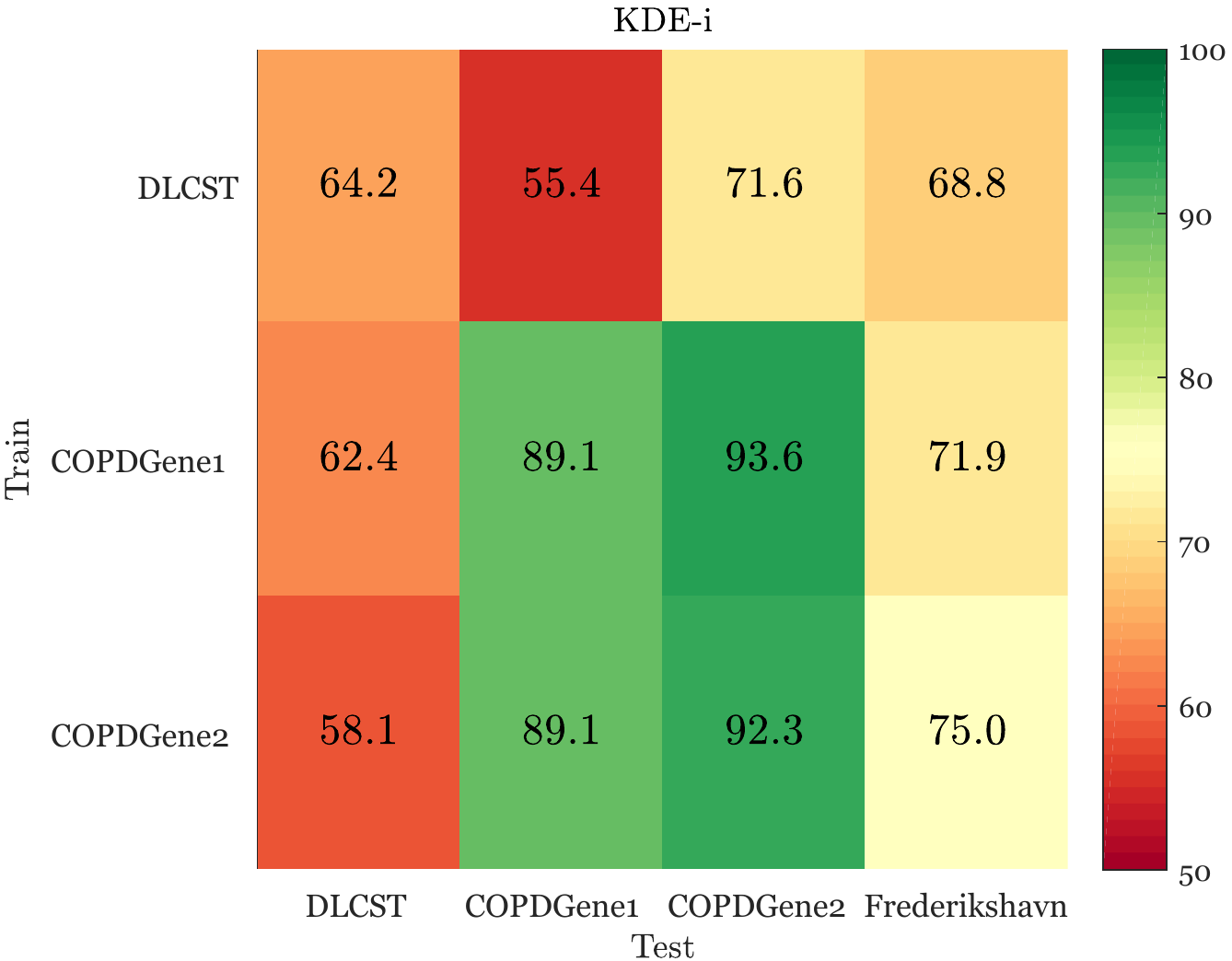}
  
    \caption{AUC $\times 100$ of SimpleMIL across datasets, without transfer, for four different feature types. Three datasets (rows) are used for training and four datasets (columns) are used for testing. Diagonal elements (for DLCST, COPDGene1 and COPDGene2) show leave-one-out performance within a single dataset.} 
    \label{fig:traintest}
\end{figure*}

Fig.~\ref{fig:traintest} shows the results of different features for the SimpleMIL logistic classifier, without using any transfer learning. In this section we summarize the results per test dataset. 

For DLCST, the best results are obtained when training within the same dataset using GSS \added{(AUC 0.790)} or GSS-t features \added{(AUC 0.779)}. The AUCs are not very high compared to those of other datasets, but they are consistent with previous results on DLCST~\cite{cheplygina2014classification,sorensen2012texture}. 

\added{For COPDGene1 and COPDGene2 we obtain much higher AUCs than for DLCST, ranging between AUC 0.850 and 0.956}. When training on one dataset and testing on the other, the performances are similar to when training within a single dataset, suggesting the protocol was well-standardized and that using a slightly different scanner did not have a large effect on the scans. In this cross-dataset scenario, all features give good results, with GSS-i being slightly better \added{(AUC 0.917 and 0.953)} than the others. However, when training on a very different dataset, DLCST, the situation changes: the best results are still provided by GSS-i \added{(AUC 0.879 and 0.859)}, but the gap between GSS-i and the other features now increases. In particular, the performance of KDE-i drops dramatically to \added{AUC 0.554 and 0.716}.

The Frederikshavn dataset also can be classified well, but the success is more dependent on the dataset and the features used, than is the case for COPDGene. The best performances on Frederikshavn are obtained with GSS-t features \added{(AUC between 0.938 and 0.953)}, followed by GSS \added{(AUC between 0.813 and 0.906)}. The two types of intensity features perform the worst, with GSS-i doing slightly better than KDE-i.

\subsection{Performance with Transfer Learning}

We now examine the performances of the importance-weighted classifiers. The performances are shown in Table~\ref{tab:allperflog} for completeness, but for better interpretation of the numbers, a summary is provided in Table~\ref{tab:summarylog}. In total we considered nine across-domain experiments. Averaged over these nine experiments, we report the AUC, the rank of each weighting method (per feature), and the rank of each feature (per weighting method). 

The average AUCs do not give a conclusive answer about whether weighting is beneficial. For both types of intensity features, weighting always improves performance, but for GSS and GSS-t weights can also deteriorate the performance slightly. The best results are obtained with the logistic weights, which improves average performance for all feature types.

The average ranks for the weights, per feature, tell a slightly different story, although here almost none of the differences are significant. For GSS, none of the weighting methods rank higher than the unweighted case. For the other features, it is always beneficial to do some form of weighting, but the best method varies per feature. In general, the differences between the ranks are quite small and not significant. The only exception is GSS-i, where \emph{s2t} has a much better rank than the other methods, and it is also the only feature for which any significant differences in ranks are found.

When comparing the ranks of the features, the differences are much larger. Now significant differences are found for each weight type. GSS features are clearly the best overall, with ranks close to 1, followed by GSS-i and GSS-t (although these differences are not significant), and KDE-i are the worst, with ranks close to 4. The last difference is significant for all weighting strategies.

\begin{table*}[ht]
\begin{center}
\caption{AUC of SimpleMIL, in percentage. In each of the nine experiments, the AUCs are compared with a DeLong test for AUCs. Per column of 4 methods, \textbf{bold}: best or not significantly worse than best different-domain method. Per row of 4 features, \underline{underline}: best or not significantly worse than best feature.}

\begin{tabular}{l | l l l l | l l l l | l l l l}
      &   gss & gss-t & gss-i & kde-i &   gss & gss-t & gss-i & kde-i &   gss & gss-t & gss-i & kde-i \\ 
     
     \hline
     
Train DLCST & \multicolumn{4}{c}{Test COPDGene1} & \multicolumn{4}{c}{Test COPDGene2} & \multicolumn{4}{c}{Test Frederikshavn} \\     
     
 \hline 

 none        & {\bf 78.4 }& {\bf 73.4 }& {\bf \underline{\smash{87.9 }}}& {\bf 55.4 }& {\bf \underline{\smash{84.7 }}}& {\bf \underline{\smash{82.7 }}}& \underline{\smash{85.9 }}& {\bf 71.6 }& {\bf \underline{\smash{90.6 }}}& {\bf \underline{\smash{95.3 }}}& {\bf \underline{\smash{75.0 }}}& {\bf \underline{\smash{68.8 }}}\\
s2t & {\bf 78.6 }& {\bf 75.2 }& {\bf \underline{\smash{89.1 }}}& 55.6 & {\bf \underline{\smash{85.8 }}}& {\bf \underline{\smash{83.8 }}}& {\bf \underline{\smash{88.5 }}}& {\bf 72.8 }& {\bf \underline{\smash{89.1 }}}& {\bf \underline{\smash{93.8 }}}& {\bf \underline{\smash{76.6 }}}& {\bf \underline{\smash{76.6 }}}\\
t2s & 77.0 & {\bf 73.2 }& \underline{\smash{86.3 }}& {\bf 57.8 }& {\bf \underline{\smash{84.0 }}}& {\bf \underline{\smash{83.0 }}}& \underline{\smash{86.0 }}& {\bf 73.5 }& {\bf \underline{\smash{90.6 }}}& {\bf \underline{\smash{95.3 }}}& \underline{\smash{76.6 }}& {\bf \underline{\smash{76.6 }}}\\
log    & {\bf 77.9 }& {\bf 73.1 }& {\bf \underline{\smash{87.4 }}}& {\bf 57.1 }& {\bf \underline{\smash{84.0 }}}& {\bf \underline{\smash{82.3 }}}& {\bf \underline{\smash{86.8 }}}& {\bf 73.3 }& {\bf \underline{\smash{93.8 }}}& {\bf \underline{\smash{96.9 }}}& {\bf \underline{\smash{75.0 }}}& {\bf \underline{\smash{78.1 }}}\\

\hline

 Train COPDGene1 & \multicolumn{4}{c}{Test DLCST} & \multicolumn{4}{c}{Test COPDGene2} & \multicolumn{4}{c}{Test Frederikshavn} \\     
 
 \hline
 
 none        & {\bf \underline{\smash{67.4 }}}& {\bf \underline{\smash{66.7 }}}& 64.8 & {\bf 62.4 }& {\bf \underline{\smash{95.6 }}}& {\bf \underline{\smash{94.1 }}}& {\bf \underline{\smash{95.3 }}}& {\bf \underline{\smash{93.6 }}}& {\bf \underline{\smash{81.3 }}}& {\bf \underline{\smash{93.8 }}}& {\bf \underline{\smash{78.1 }}}& {\bf \underline{\smash{71.9 }}}\\
s2t & {\bf \underline{\smash{67.0 }}}& \underline{\smash{65.6 }}& {\bf \underline{\smash{65.8 }}}& {\bf 62.1 }& {\bf \underline{\smash{95.7 }}}& {\bf \underline{\smash{95.0 }}}& {\bf \underline{\smash{95.0 }}}& {\bf \underline{\smash{93.4 }}}& \underline{\smash{81.3 }}& {\bf \underline{\smash{95.3 }}}& {\bf \underline{\smash{79.7 }}}& \underline{\smash{71.9 }}\\
t2s & {\bf \underline{\smash{67.2 }}}& {\bf \underline{\smash{66.3 }}}& {\bf 65.8 }& {\bf 62.0 }& {\bf \underline{\smash{96.2 }}}& {\bf \underline{\smash{94.5 }}}& {\bf \underline{\smash{95.8 }}}& {\bf \underline{\smash{93.5 }}}& {\bf \underline{\smash{79.7 }}}& {\bf \underline{\smash{92.2 }}}& \underline{\smash{79.7 }}& \underline{\smash{71.9 }}\\
log    & \underline{\smash{67.0 }}& {\bf \underline{\smash{66.7 }}}& 65.3 & {\bf 62.1 }& {\bf \underline{\smash{95.5 }}}& {\bf \underline{\smash{94.6 }}}& {\bf \underline{\smash{95.2 }}}& {\bf \underline{\smash{93.4 }}}& \underline{\smash{81.3 }}& {\bf \underline{\smash{96.9 }}}& \underline{\smash{79.7 }}& 71.9 \\

 \hline 
 
 Train COPDGene2 & \multicolumn{4}{c}{Test DLCST} & \multicolumn{4}{c}{Test COPDGene1} & \multicolumn{4}{c}{Test Frederikshavn} \\

 \hline 
none        & \underline{\smash{66.9 }}& \underline{\smash{66.6 }}& {\bf \underline{\smash{65.4 }}}& 58.1 & {\bf \underline{\smash{90.8 }}}& {\bf \underline{\smash{86.1 }}}& {\bf \underline{\smash{91.7 }}}& {\bf \underline{\smash{89.1 }}}& {\bf \underline{\smash{87.5 }}}& {\bf \underline{\smash{93.8 }}}& {\bf \underline{\smash{73.4 }}}& {\bf \underline{\smash{75.0 }}}\\
s2t & \underline{\smash{65.4 }}& \underline{\smash{62.7 }}& {\bf \underline{\smash{65.6 }}}& {\bf 61.7 }& {\bf \underline{\smash{89.9 }}}& {\bf 85.6 }& {\bf \underline{\smash{91.9 }}}& {\bf \underline{\smash{88.9 }}}& {\bf \underline{\smash{85.9 }}}& {\bf \underline{\smash{93.8 }}}& {\bf \underline{\smash{76.6 }}}& {\bf \underline{\smash{76.6 }}}\\
t2s & {\bf \underline{\smash{67.9 }}}& \underline{\smash{66.3 }}& 65.2 & {\bf 61.6 }& {\bf \underline{\smash{90.7 }}}& {\bf \underline{\smash{86.2 }}}& {\bf \underline{\smash{91.5 }}}& {\bf \underline{\smash{89.4 }}}& {\bf \underline{\smash{87.5 }}}& {\bf \underline{\smash{96.9 }}}& {\bf \underline{\smash{75.0 }}}& {\bf \underline{\smash{78.1 }}}\\
log    & {\bf \underline{\smash{68.4 }}}& {\bf \underline{\smash{67.7 }}}& {\bf 65.5 }& {\bf 61.6 }& {\bf \underline{\smash{90.7 }}}& {\bf \underline{\smash{86.5 }}}& {\bf \underline{\smash{91.6 }}}& {\bf \underline{\smash{89.4 }}}& {\bf \underline{\smash{89.1 }}}& {\bf \underline{\smash{95.3 }}}& {\bf \underline{\smash{75.0 }}}& \underline{\smash{78.1 }}\\

\end{tabular}

\label{tab:allperflog}
\end{center}
\end{table*}

\begin{table}
\begin{center}

\caption{Top: Average AUC, in percentage, over nine transfer experiments. Best weighting method is in bold, best feature is underlined. Middle: Ranks of each weight type (1=best, 4=worst), compare per column. Best weight, or weights that are not significantly worse (Friedman/Nemenyi test, critical difference = 1.56) are in bold. Bottom: Ranks of each feature type (1=best, 4=worst), compare per row. Best feature, or features that are not significantly worse (Friedman test, critical difference = 1.56) are underlined.}

\begin{tabular}{l | l l l l  }

 & gss & gss-t & gss-i & kde-i \\
 \hline
 
none        & 82.6 &\underline{\smash{83.6}} & 79.7 & 71.8 \\
s2t & 82.1 &\underline{\smash{83.4}} &  {\bf 81.0 } & 73.3 \\
t2s & 82.3 &\underline{\smash{83.8}} & 80.2 & 73.8 \\
log    &  {\bf 83.1} &{\bf\underline{\smash{84.4}}} & 80.2 &  {\bf 73.9} \\

\hline

none        & {\bf 2.11} & {\bf 2.78} & 3.17 & {\bf 3.06} \\
s2t & {\bf 2.78} & {\bf 2.72} & {\bf 1.50 } & {\bf 2.78} \\
t2s & {\bf 2.78} & {\bf 2.61} & {\bf 2.67 }& {\bf 2.06} \\
log    & {\bf 2.33} & {\bf 1.89} & {\bf 2.67} & {\bf 2.11} \\

\hline

none        & \underline{\smash{1.67}} & \underline{\smash{2.22}} & \underline{\smash{2.33}} & 3.78 \\
s2t & \underline{\smash{1.78}} & \underline{\smash{2.44}} & \underline{\smash{2.00}} & 3.78 \\
t2s & \underline{\smash{1.72}} & \underline{\smash{2.22}} & \underline{\smash{2.33}} & 3.72 \\
log    & \underline{\smash{1.67}} & \underline{\smash{2.22}} & \underline{\smash{2.44}} & 3.67 \\

\end{tabular}

\label{tab:summarylog}
\end{center}
\end{table}

%

\section{Discussion}\label{sec:discussion}

The main findings from the previous section are: (i) there are large differences between datasets, (ii) there are large differences between features, and (iii) weighting, in particular with logistic classifier-based weights, can improve performance. In this section we discuss these results in more detail. We then discuss limitations of our method, and provide some recommendations for classification of COPD in multi-center datasets.

\subsection{Datasets}
\begin{figure*}
    \centering
    \includegraphics[width=0.9\textwidth]{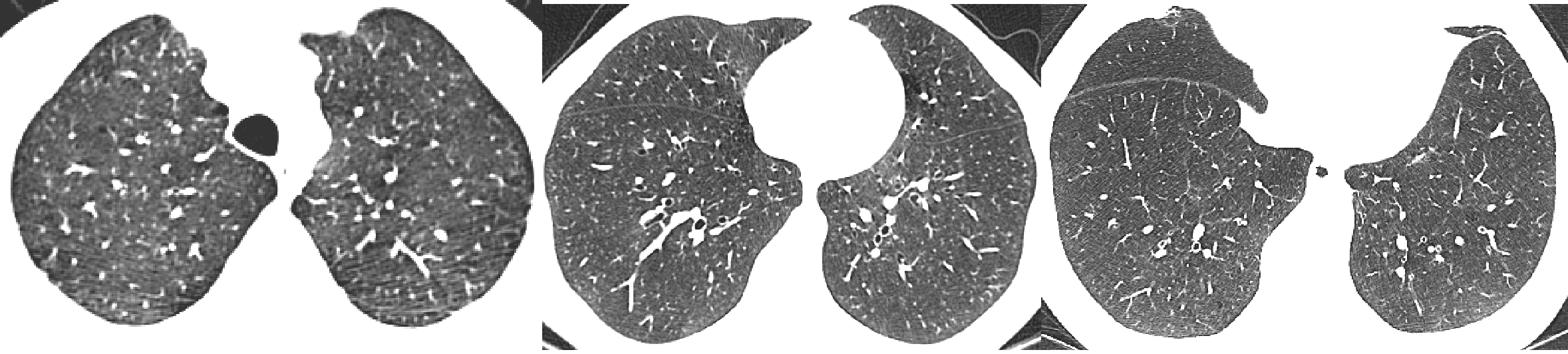}
    \caption{\added{Examples of slices from the DLCST, COPDGene1 and Frederikshavn datasets.}}
    \label{fig:scans}
\end{figure*}

\added{The datasets differ in several ways. First of all, the acquisition parameters lead to differences in the appearances of the scans. This is illustrated in Fig.\ref{fig:scans}. The acquisition parameters are the most different for the DLCST data, which can be also seen in the visual appearance in the images. This can partially explain the lower performances, especially with intensity features, when training on the DLCST dataset.}

\added{Another important difference is the distribution of COPD severity, which affects how high the performances can become in general}. For example, DLCST is more difficult to classify than the other datasets. The highest AUC for DLCST is 0.79 (when training on the same domain), whereas the AUCs for the other datasets are often higher than 0.9. This difference can be explained by the differences in COPD severity between datasets. DLCST contains many cases of mild COPD, which can be easily misclassified as healthy subjects. The other datasets contain more severe cases of COPD. This is supported by the fact that, if we remove the GOLD 2-4 subjects from the 
COPDGene datasets, the AUC decreases to around 0.8 for the best features. 

The datasets have different sizes, which can also affect the results of the classifiers. When the training and test data are from the same or similar domain, the training dataset should be sufficiently large to describe all possible variations. As a result, when testing on COPDGene2, it is actually better to train on COPDGene1 (which has similar scans, but is larger than COPDGene2), than to do same-domain training on COPDGene2. Another example is Frederikshavn: since both DLCST and the COPDGene datasets are rather dissimilar, the larger DLCST training data tends to give better results. As such, it would be interesting to compare results of different methods, when sampling the same number of training scans from each dataset.

\subsection{Features}

Our results show that intensity is not always a robust choice of features when classifying across domains. Gaussian scale space features, which combine intensity and texture components, had higher performances overall, and in some cases, the intensity components could even deteriorate the performance.  
These findings are interesting with respect to previous results from the literature. On a task of classifying ROIs within a single domain, \cite{sorensen2010quantitative} showed that local binary pattern (LBP) texture features combined with intensity features can give good classification performance. However, \cite{mendoza2012emphysema} showed that intensity features alone performed better than a different implementation of LBP in across domain classification. 

We note that there are several differences between \cite{mendoza2012emphysema} and the current study. We focus on weakly-supervised classification of entire chest CT scans, whereas \cite{mendoza2012emphysema} deals with a multi-class ROI classification problem.  Furthermore, in our transfer learning experiments the training and test domains are disjoint, i.e., the classifier does not have access to any labeled data from the test domain. On the other hand, in~\cite{mendoza2012emphysema} scans from the same domain are present in the training set. Combined with their use of the nearest neighbor classifier, this could enable intensity features to perform well even if intensities are different across domains. A further difference is that to avoid overfitting, we reduced the dimensionality of the intensity representation.

\subsection{\added{Classifier}}\label{sec:discussion_classifier}

\added{We chose SimpleMIL with logistic classifier and the average assumption due to its good performance on a similar problem~\cite{cheplygina2014classification}. This classifier assumes that all instances contribute to the bag label, i.e., for a subject to have COPD, the ROIs must on average be classified as having disease patterns. This reflects the idea that COPD is a diffuse disease, affecting large parts of the lung rather than small isolated regions~\cite{muller2002chronic}.}

\added{An alternative assumption is the traditional, "noisy-or" MIL assumption~\cite{dietterich1997solving}, which is defined as follows:}

\begin{equation}\label{eq:noisyor}
\frac{p(y=1|Z_i)}{p(y=-1|Z_i)} = \frac{1 - \prod_{j=1}^{n_i} (1-p(y_{ij}=1|\mathbf{z}_{ij}))}{\prod_{j=1}^{n_i} p(y_{ij}=-1|\mathbf{z}_{ij})}
\end{equation}

\added{In this case, for a subject to have COPD, it is sufficient that only one ROI which has a high probability of having disease patterns, i.e. a value of $p(y_{ij}=1|\mathbf{z}_{ij}$ close to 1. Although this assumption is intuitive for some computer-aided diagnosis applications, in practice it is less robust than the average assumption because there is class overlap between positive and negative instances, and individual instances can be easily misclassified. Relying only the label of the most positive instance therefore leads to bag-level errors.} 

\added{We did post-hoc experiments to investigate how replacing the average assumption with the noisy-or assumption would affect the results. For this we performed only the baseline experiments without weighting, for the GSS features, within and across datasets (i.e. the numbers that can be seen in the top left of Figure~\ref{fig:traintest}). In Table~\ref{tab:noisyor}  we again present these performances, next to the performances of using the classifier with the noisy-or assumption. Here we can see that overall, the average assumption outperforms the noisy-or assumptions, with a few exceptions. With noisy-or, in particular the performances across dissimilar datasets deteriorate, as can be seen when a combination of DLCST and one of the COPDGene datasets is used.}

\begin{table}

\begin{center}

\caption{\added{AUC ($\times 100$) of SimpleMIL across datasets, without transfer, for GSS features and two different assumptions: average assumption used in this paper (top) and noisy-or assumption (bottom).}}

\begin{tabular}{l | l l l l  }

Test $\rightarrow$ & DLCST & COPDGene1 & COPDGene2 & Frederikshavn \\
\hline
Train $\downarrow$ & \multicolumn{4}{c}{Average assumption}\\

DLCST &    79.0   &  78.4  &  84.7 &   90.6\\
COPDGene1 &   67.4  & 88.4 &  95.6 &   81.3\\
COPDGene2 &   66.9  &  90.8 &  93.1 &    87.5\\

& \multicolumn{4}{c}{Noisy-or assumption}\\

DLCST &     74.8 &    62.9 &    61.0 &   92.2 \\
COPDGene1 & 50.7 &    76.2 &    96.0 &   77.3  \\
COPDGene2 & 50.0 &   84.6  &   81.1 &    87.5 \\

\end{tabular}
\end{center}
\label{tab:noisyor}

\end{table}



\subsection{Weights}

Weighting can improve performance across domains, but does not guarantee improved performance. In our experiments, no weighting method was always (for each dataset combination and feature type) better than the unweighted baseline. However, on average the logistic classifier-based weights performed quite well. The logistic weights had the highest average performance on each of the four feature types, and the highest rank on three out of four features, although the differences were not significant. 

The small difference between \emph{s2t} and \emph{t2s}, the different ways in which source and target bags can be compared, is interesting. In a study of brain tissue segmentation across scanners~\cite{cheplygina2016asymmetric}, weighting trained \emph{classifiers} based on the \emph{t2s} distance was more effective than weighting them based on the \emph{s2t} distance. We thus hypothesized that \emph{t2s} might also be a better strategy for weighting training samples, but our results show that this is not the case.  

\begin{figure}
    \centering
    \includegraphics[width=0.44\textwidth]{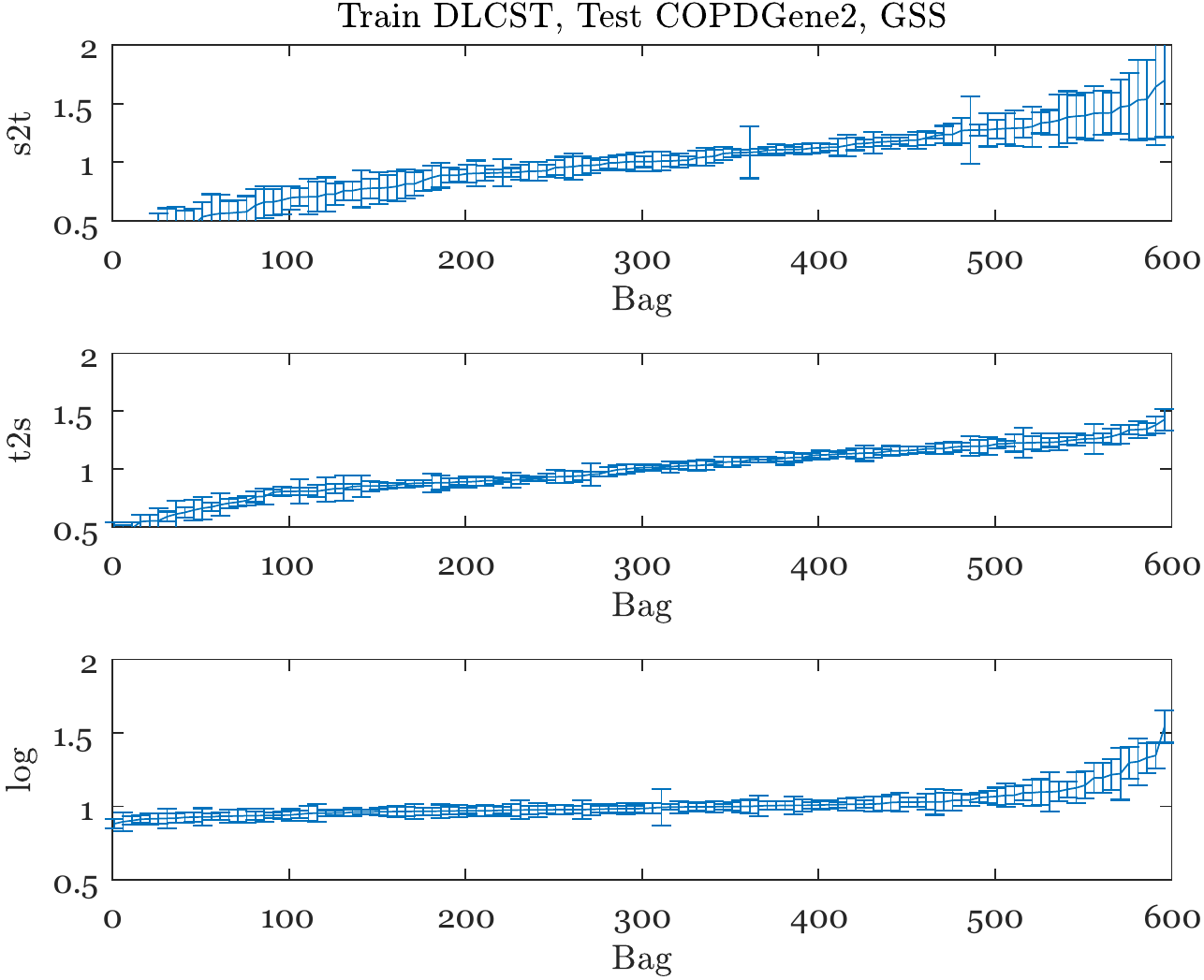}
    \includegraphics[width=0.44\textwidth]{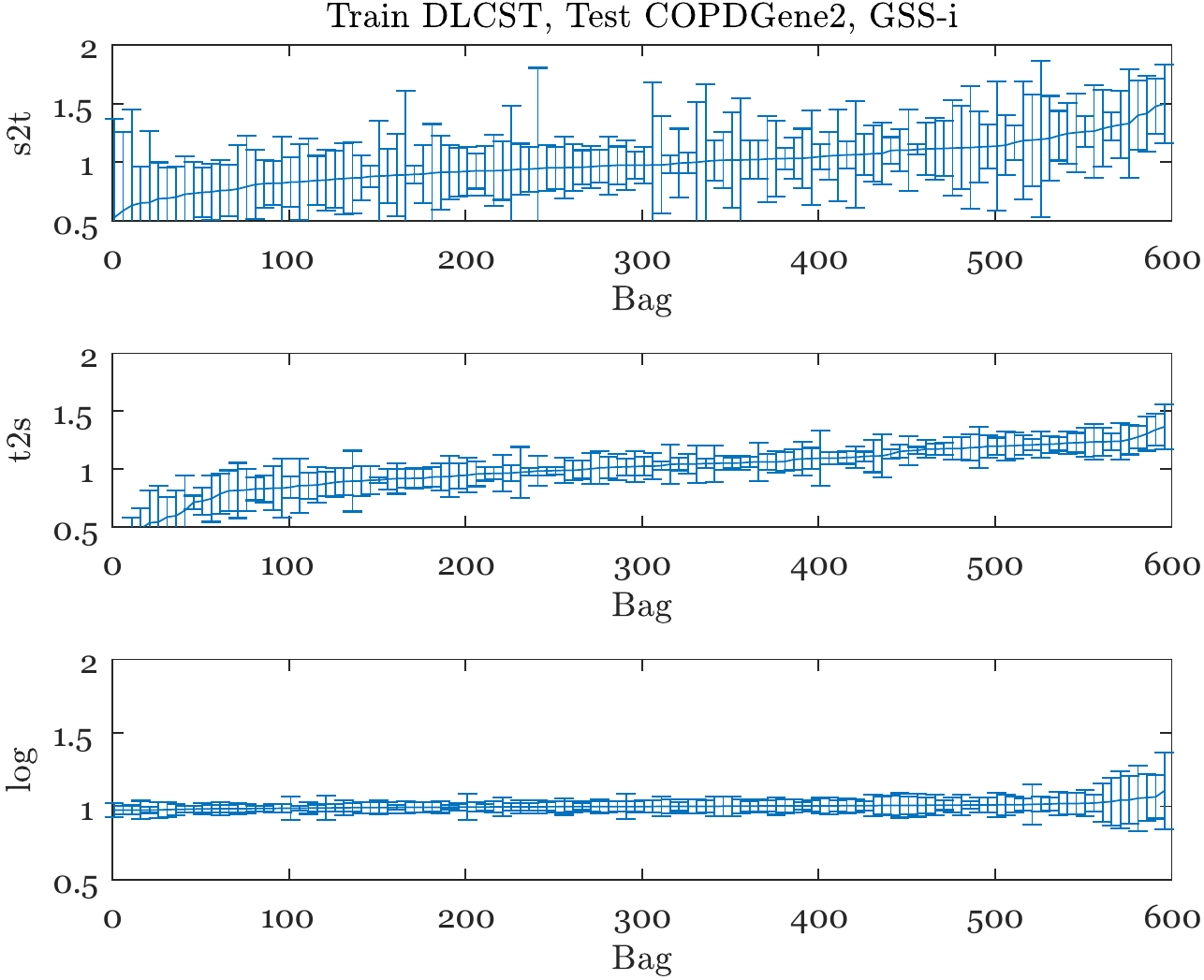}
   
    \caption{Distribution of weights when training on DLCST and testing on COPDGene2, for the GSS (top three plots) and GSS-i (bottom three plots) features. The mean and standard deviation of the weight per training bag is shown, for every 5th (due to the large number of bags) training bag in DLCST. The training bags are sorted by average weight for better visualization, there is therefore no correspondence between different x-axes.} 
     \label{fig:weights}
\end{figure}

To further understand the differences between the weights, we looked at the weights assigned to each training bag. In Fig.~\ref{fig:weights} we show the weights when training on DLCST and testing on COPDGene2 for two of the feature types: GSS with 320 features and GSS-i with 40 features. In each case, we first find the mean and the standard deviation of the weights, assigned to each training bag. We then sort the training bags by their mean weight, and plot the mean and the standard deviations with error bars.

Per training bag, the distance-based weights have a higher variance than the logistic weights. Furthermore, with distance-based weights, the distributions are more steep, i.e. more training bags have a very low, or a very high average weight. Setting many weights (almost) to zero, as is the case for the distance-based weights, effectively decreases the sample size, possibly resulting in lower performance. 

One of the reasons for this behavior is the way that the weights are scaled. With the logistic weights, the exponential function provides a more natural scaling of the weights. For example, if all the source bags are similar to the target bag, they will all receive similar weights. The scaling we apply for the distance-based weights is more ``artificial'', because the most similar bag is assumed to have weight 1, and the least similar bag is assumed to have weight 0. Furthermore, logistic weights are based on all the source bags, i.e., they are assigned by a classifier trained to distinguish the target bag from all the source bags. On the other hand, the distance-based weights are based only on the distance between the target bag and each individual source bag, which leads to noise. 

In Fig.~\ref{fig:weights} we also see that the differences between the weight types are much larger for GSS-i. This is consistent with the fact that we observe smaller differences in AUC performances for GSS. This might be caused by the differences in dimensionality: in higher dimensions, distances become more and more similar, reducing the differences in the weights. The logistic weights are the most robust to the difference in dimensionality.

We now focus on the logistic weights, as these weights perform better on average. Examining their effect on different combinations of source and target datasets, we see that they have the most benefit when the datasets have different scan protocols. Using logistic weights when training on COPDGene1 and testing on COPDGene2 and vice versa has only small improvements, or even deteriorates the performance. This suggests that both the marginal distributions $p(\mathbf{x})$ and the labeling functions $p(y|\mathbf{x})$ and of these datasets are very similar, due to a similar distribution of subjects and the same scanning protocol.

\subsection{Limitations}\label{sec:limitations}

\subsubsection{Intensity normalization}
\added{Normalization has been shown to reduce differences between quantitative emphysema~\cite{mol2010correction} or air-trapping~\cite{choi2014improved} measures from different scanners, and to improve correlations between emphysema and spirometry in scans reconstructed with different kernels\cite{gallardo2016normalizing}}. However, as these studies use different strategies, there is no widely accepted way to perform normalization in chest CT and in theory, Hounsfield units should be comparable between scans, we initially did not perform normalization in this study.

\added{We later conducted a post-hoc investigation of the effect of intensity normalization on our results. We normalized intensity by fitting a Gaussian to voxels inside the trachea, and shifted the intensities such that the peak of the Gaussian was at -1000 HU. This is the approach described in~\cite{mol2010correction}, a related approach is taken by \cite{choi2014improved}.} 

\added{We then extracted features, and performed experiments for the nine combinations of training and test datasets. In these experiments we observed the same performances for GSS-t (since a translation of intensities does not affect the gradient), and similar or slightly lower performances for GSS and GSS-i. For KDE-i, the performances became much better when training on DLCST and testing on COPDGene1 and COPDGene2, but slightly decreased in all other cases. We attribute these results to the binning of the intensities. Although the intensities are more comparable after normalization, which leads to some improved results, we use the same bins as in the experiments without normalization, which may not be optimal anymore.} 

\added{The experiments with intensity normalization are summarized in Table \ref{tab:summaryintnorm}. We see that the average AUCs are slightly lower than in Table \ref{tab:summarylog} for all features except GSS-t. Weighting still generally has a benefit, but the ranks of the weights are closer to each other, and therefore the differences in weights are not significant. However, the logistic weights rank better than the unweighted methods for all features used.} 

\added{The ranking of the features is still the same as in the scenario without normalization. GSS has the best rank, followed closely by GSS-t, then by GSS-i and finally KDE-i. However, the differences between the best and worst ranks are also smaller, which means the differences are not significant.} 

\added{Overall, we conclude that intensity normalization by trachea air is a useful tool for making intensities more comparable. However, a weighting strategy can still be beneficial, although the differences are less pronounced than if no normalization is used. Finally, Gaussian texture-only features still remain a robust choice of features which remove the need for intensity normalization. This also removes the need to perform segmentation of the trachea as a preprocessing step.}

\begin{table}
\begin{center}

\caption{\added{Top: Average AUC, in percentage, over nine transfer experiments. Best weighting method is in bold, best feature is underlined. Middle: Ranks of each weight type (1=best, 4=worst), compare per column. The differences are not significant according to the Friedman/Nemenyi test. Bottom: Ranks of each feature type (1=best, 4=worst), compare per row. The differences are not significant according to the Friedman/Nemenyi test.}}

\begin{tabular}{l | l l l l  }

 & gss & gss-t & gss-i & kde-i \\
 \hline
 none        & 78.0 & \underline{\smash{ 83.6}} & 73.2 & 70.0 \\
s2t & {\bf  78.5} & \underline{\smash{ 83.4}} & {\bf 76.4} & 70.7 \\
t2s & 78.0 & \underline{\smash{ 83.8}} & 73.5 & {\bf 71.2} \\
log    & {\bf  78.5} & {\bf \underline{\smash{ 84.4}}} & 74.1 & 71.1 \\

\hline

none        & 2.78 & 2.78 & 2.94 & 2.72 \\
s2t & 2.39 & 2.72 & 1.67 & 2.83 \\
t2s & 2.33 & 2.61 & 3.00 & 2.22 \\
log    & 2.50 & 1.89 & 2.39 & 2.22 \\

\hline
none        & 1.89 & 2.00 & 2.89 & 3.22 \\
s2t & 2.00 & 2.11 & 2.44 & 3.44 \\
t2s & 1.89 & 2.00 & 2.83 & 3.28 \\
log    & 1.89 & 2.11 & 2.67 & 3.33 \\

\end{tabular}

\label{tab:summaryintnorm}
\end{center}
\end{table}

\subsubsection{Binary classification} \label{sec:binary_classification}

\added{We considered a binary classification problem - non-COPD (GOLD 0) and COPD (GOLD 1-4) in our experiments. This is a limitation since it could be argued that identifying the early stages of disease is more difficult, but more relevant clinically. Since our classifier outputs posterior probabilities, we could use the posterior probability of a subject having COPD (i.e. $p(y=1|X))$ as an indication of COPD severity as expressed by the GOLD grade. We use the Spearman correlation between the posterior probability and the GOLD grade to investigate this, for baseline experiments with GSS and KDE-i features. The correlation coefficients are presented in Table \ref{tab:corrcoef}.}

\added{Overall we observe moderate and strong correlations, in particular when GSS features are used. The difference between GSS and KDE-i is particularly pronounced when training on DLCST and testing on DLCST or COPDGene, where the correlations decrease from moderate for GSS to weak or very weak for KDE-i. The correlations for Frederikshavn are often close to zero, except when training on DLCST and testing, where the correlation is moderate. The very weak correlations are not significant due to the small size of the dataset.}

\begin{table*}
\begin{center}
\caption{Spearman correlation between the GOLD value and $p(y=1|X)$, the posterior probability that a subject has COPD, given by the classifier. Significant correlations are in bold.}
\begin{tabular}{l | r r r r | r r r r}
         
Test $\rightarrow$         & DLCST & COPDGene1 & COPDGene2 & Frederikshavn & DLCST & COPDGene1 & COPDGene2 & Frederikshavn \\ 
Train $\downarrow$ & \multicolumn{4}{c}{GSS features} &   \multicolumn{4}{c}{KDE-i features}\\

 \hline 
DLCST     &\bf 0.49 &\bf 0.53 &\bf 0.59 & 0.09 &\bf 0.24 & 0.06 & 0.30 & \bf 0.55 \\
COPDGene1 &\bf 0.31 &\bf 0.74 &\bf 0.85 & 0.02 &\bf 0.22 &\bf 0.73 &\bf 0.81 & -0.08 \\
COPDGene2 &\bf 0.30 &\bf 0.77 &\bf 0.83 & 0.00 &\bf 0.12 &\bf 0.69 &\bf 0.83 & -0.02 \\
\end{tabular}
\end{center}
\label{tab:corrcoef}
\end{table*}

\subsection{Recommendations}

Based on our observations, we offer some advice to researchers who might be faced with classification problems involving scans from different scanners. 

\begin{itemize}

    \item Adaptive histograms of multi-scale Gaussian derivatives are a robust choice of features. Although originally this specific filterbank was used for classifying ROIs~\cite{sorensen2010quantitative} and later classifying DLCST scans~\cite{sorensen2012texture}, we did not need any modifications to successfully apply them to independent datasets. 
    
    \item If using intensity histogram features, adaptive binning is a good way to focus on the more informative intensity ranges, while keeping the dimensionality low. Reducing the dimensionality in KDE only reduces the number of bins, but does not consider their information content. As such, bins in informative intensity ranges become too wide, reducing the classification performance.
    
    \item Randomly sampled ROIs together with a SimpleMIL logistic classifier that uses the averaging rule is a good starting point for distinguishing COPD from non-COPD scans, achieving at most 79.0 (DLCST), 91.7 (COPDGene1), 95.6 (COPDGene2) and 95.3 (Frederikshavn) AUC, in \%. 
    
    \item Importance weighting appears not to be needed when the same cohort and only a slightly different scan protocol are used, such as with the COPDGene datasets. 
    
    \item Importance weights based on a logistic classifier trained to discriminate between source data and target data, are a good starting point. These weights gave the best results overall, eliminate the scaling problem, and were much faster to compute (2 seconds per test image) than the distance-based weights (2 minutes per test image) in this study.
    
\end{itemize}

\section{Conclusions}

We presented a method for COPD classification using a chest CT scan which generalizes well to datasets acquired at different sites and scanners. Our method is based on Gaussian scale-space features and multiple instance learning with a weighted logistic classifier. Weighting the training samples according to their similarity to the target data could further improve the performance, demonstrating the potential benefit of transfer learning techniques in this problem. Transfer learning methods beyond instance-transfer approaches could be interesting in the future. To this end, upon acceptance of the paper we will publicly release the DLCST and Frederikshavn datasets to encourage more investigation into transfer learning methods in medical imaging. We believe that developing methods that are robust across domains is an important step for adoption of automatic classification techniques in clinical studies and clinical practice.

\section*{Acknowledgements}

This research was performed as part of the research project ``Transfer learning in biomedical image analysis'' which is financed by the Netherlands Organization for Scientific Research (NWO) grant no. 639.022.010. We thank Morten Vuust MD (Department of Diagnostic Imaging, Vendsyssel Hospital, Frederikshavn) and  Ulla M{\o}ller Weinreich (Department of Pulmonology Medicine and Clinical Institute, Aalborg University Hospital) for their assistance in the acquisition of the data used in this study. We thank the Danish Council for Independent Research for partial support of this research. We thank Wouter Kouw (Pattern Recognition Laboratory, Delft University of Technology) for discussions on logistic weights.

\bibliographystyle{IEEEtran}	%
\bibliography{refs}	

\end{document}